\documentclass[letterpaper]{article} 
\usepackage{aaai2026}  
\usepackage{times}  
\usepackage{helvet}  
\usepackage{courier}  
\usepackage[hyphens]{url}  
\usepackage{graphicx} 
\usepackage{natbib}  
\usepackage{caption} 
\usepackage{algorithm}
\usepackage{algorithmic}
\usepackage{booktabs, makecell}
\newtheorem{myDef}{Definition}

\usepackage{newfloat}
\usepackage{listings}
\usepackage{multirow}
\usepackage{amsmath,amssymb,amsfonts}
\usepackage{xspace}
\newcommand{\oursys}{PAF-Net\xspace}
\usepackage{newfloat}
\usepackage{listings}
\DeclareCaptionStyle{ruled}{labelfont=normalfont,labelsep=colon,strut=off} 
\floatstyle{ruled}
\newfloat{listing}{tb}{lst}{}
\floatname{listing}{Listing}
 \nocopyright 

\title{PAF-Net: Phase-Aligned Frequency Decoupling Network for Multi-Process Manufacturing Quality Prediction}
\author{
    Yang Luo\textsuperscript{\rm 1},
    Haoyang Luan\textsuperscript{\rm 1},
    Haoyun Pan\textsuperscript{\rm 1},
    Yongquan Jia\textsuperscript{\rm 2},
    Xiaofeng Gao\textsuperscript{\rm 1}\thanks{Xiaofeng Gao is the corresponding author},
    Guihai Chen\textsuperscript{\rm 1}
}
\affiliations{
    \textsuperscript{\rm 1} School of Computer Science, Shanghai Jiao Tong University, Shanghai 200240, China\\
    Email: \{floating-dream, stevenluan, adam\_pan\}@sjtu.edu.cn; \{gao-xf, gchen\}@cs.sjtu.edu.cn\\
    \textsuperscript{\rm 2}NIO Inc.  Hefei, Anhui, China\\
}

\usepackage{bibentry}

\begin{document}

\maketitle

\begin{abstract}
Accurate quality prediction in multi-process manufacturing is critical for industrial efficiency but hindered by three core challenges: time-lagged process interactions, overlapping operations with mixed periodicity, and inter-process dependencies in shared frequency bands. To address these, we propose PAF-Net, a frequency-decoupled time series prediction framework  with three key innovations: (1) A phase-correlation alignment method guided by frequency-domain energy to synchronize time-lagged quality series, resolving temporal misalignment. (2) A frequency-independent patch attention mechanism paired with Discrete Cosine Transform (DCT) decomposition to capture heterogeneous operational features within individual series. (3) A frequency-decoupled cross-attention module that suppresses noise from irrelevant frequencies, focusing exclusively on meaningful dependencies within shared bands. Experiments on $4$ real-world datasets demonstrate PAF-Net’s superiority. It outperforms 10 well-acknowledged baselines by 7.06\% lower MSE and 3.88\% lower MAE.  Our Code is available at https://github.com/StevenLuan904/PAF-Net-Official.
\end{abstract}

%

\section{Introduction}
As global initiatives like Industry 5.0 prioritize AI-driven quality assurance, the smart manufacturing market is anticipated to grow from USD $233.33$ billion in $2024$ to $479.17$ billion by $2029$\footnote{https://www.marketsandmarkets.com/Market-Reports/smart-manufacturing-market-105448439.html}. Within this expanding sector, quality prediction stands as a critical component. By enabling anomaly detection and timely adjustments to manufacturing process, it enhances product quality and operational efficiency, avoiding the economic losses associated with physical destructive testing. These combined benefits solidify its role as a cornerstone of competitiveness in modern manufacturing.

\begin{figure*}[htbp]
    \centering
    \includegraphics[width=\linewidth]{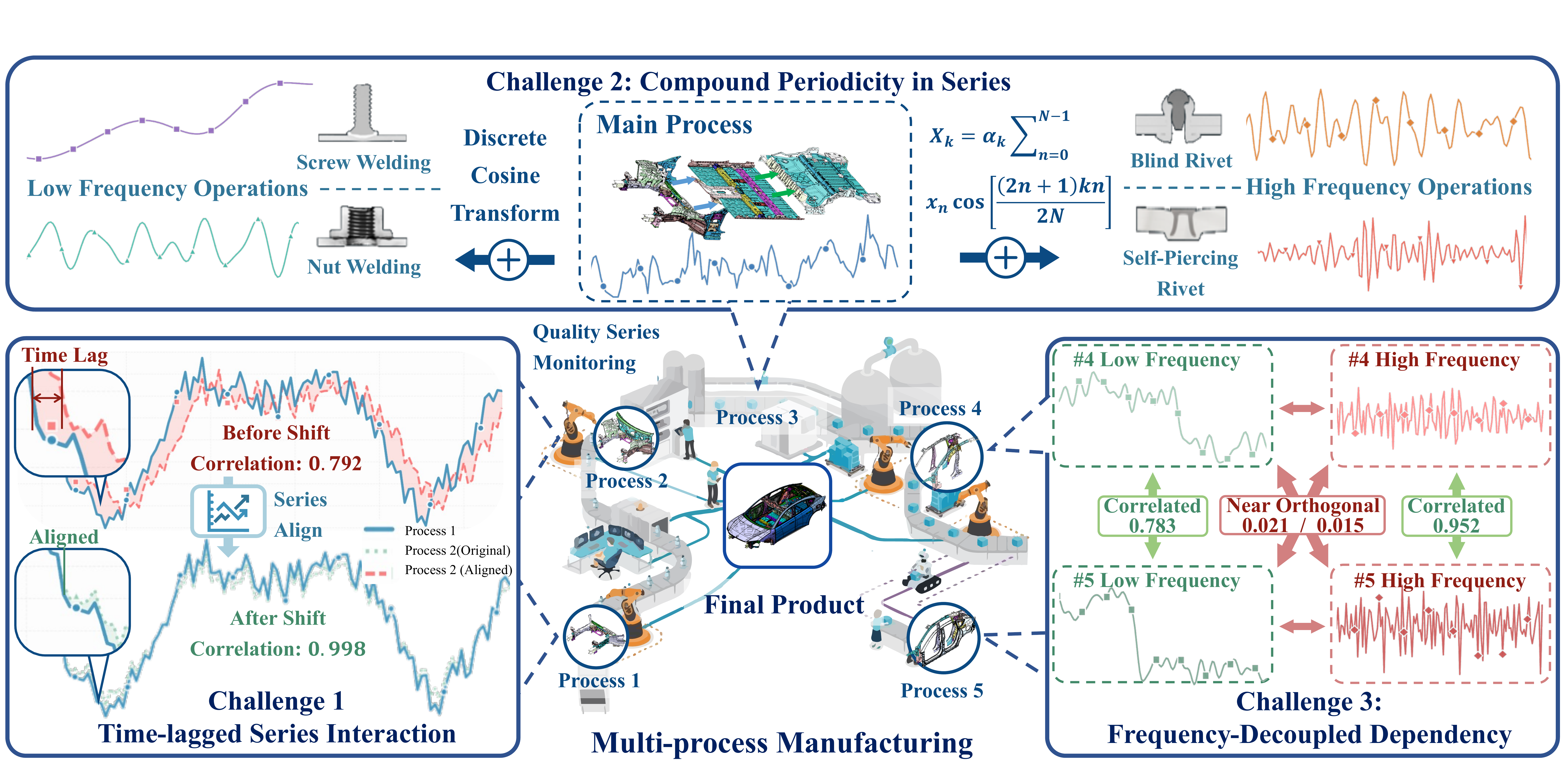}
    \caption{Schematic of Real-World Manufacturing. The middle part shows the vehicle production chain with multiple processes. Three subgraphs illustrate identified challenges. Correlations in Challenges 1 and 3 are calculated by the Pearson Coefficient.}
    \label{fig:intro}
\end{figure*}
Fig.~\ref{fig:intro} presents a multi-process production scenario, where each process is equipped with sensors to record real-time quality metrics. Historical quality metrics from different processes are aggregated into time series, forming a multivariate problem. The objective of quality prediction is to leverage these cross-process temporal patterns to forecast each process’s quality metric, enabling proactive anomaly detection and intervention. This task naturally lends itself to a multivariate time-series forecasting framework, which, however, presents distinct challenges, including:

\textbf{Challenge 1: Time-Lagged Series Interaction.} 
Manufacturing processes exhibit inherent sequential continuity, where upstream operations influence downstream quality metrics after a latent delay. This temporal lag disrupts the alignment of sequential dependencies, as effects manifest asynchronously in downstream series. Such misalignment obscures causal relationships between process stages, hindering accurate interaction modeling and quality prediction.


\textbf{Challenge 2: Compound Periodicity in Series.} 
A manufacturing process involves overlapping operations with distinct periodicities, such as low-frequency wielding and high-frequency Rivet Process. These heterogeneous periodic components interact, generating non-stationary patterns in quality time series. The coexistence of multiple frequencies and phases within a series complicates isolating individual operational impacts, hindering accurate prediction.


\textbf{Challenge 3:  Frequency-Decoupled Dependency.}  
Interactions between series from different processes are frequency-dependent, with meaningful correlations concentrated in shared frequency bands. The near orthogonal nature of operational signals at distinct frequencies renders cross-frequency dependencies inherently weak and less informative. To accurately model inter-series relationships, frequency-decoupled analysis is critical to examine aligned frequency components at the operational level, enabling a precise characterization of functional dependencies.

Existing multi-process quality prediction methods \cite{EAAI1, EAAI2, TC1} frame each process series as a variable in multivariate time-series frameworks. To address inter-series time lags, methods like \cite{LIFT, Timelag2, Timelag3} align series via temporal similarity metrics but overlook phase correlations across periodic frequencies, leading to suboptimal lag mitigation in periodic manufacturing \textbf{(Challenge 1)}. While recent frequency-domain studies \cite{FITS, PDF, AAAI3} effectively decompose series into periodic components, they aggregate heterogeneous frequencies, obscuring individual operational impacts on quality \textbf{(Challenge 2)}. For inter-series dependencies, graph networks \cite{MTGNN, STAMP, AAAI1} and cross-attention models \cite{Crossformer, iTransformer} are commonly used but fail to decouple compound frequencies, sacrificing operational-level physical dependencies \textbf{(Challenge 3)}. These limitations highlight critical research gaps in multi-process manufacturing quality prediction.


 To systematically address these challenges, we propose Frequency-Decoupled Network with Phase-Correlation Alignment for Multi-Process Manufacturing Quality Prediction (\oursys).  For multi-process quality series input, we first design a phase-correlation alignment method based on frequency-domain energy analysis to mitigate time-lag effects by aligning cross-process temporal patterns (\textbf{Challenge 1}). For each process series, we employ Discrete Cosine Transform (DCT) to decompose compound periodic operations into distinct frequency bands, followed by a frequency-independent patch attention mechanism to capture heterogeneous operational features at different frequencies (\textbf{Challenge 2}). Finally, we propose a frequency-decoupled cross-attention module designed to explicitly capture inter-series dependencies within shared frequency bands while suppressing cross-frequency interactions that are approximately orthogonal. The design mitigates overfitting risks and redefines dependency modeling in manufacturing through a frequency-alignment lens (\textbf{Challenge 3}).  Our contributions are summarized as follows:

\begin{itemize}
    \item We propose the phase-correlation alignment method that leverages frequency energy analysis to automatically synchronize time-lagged multi-process quality series.
    
    \item We decompose series into distinct frequency bands by DCT decomposition and design frequency-specific patch attention to model compound periodic operations.
    
    \item  We design a frequency-aligned attention module to explicitly series dependencies among same-frequency components while suppressing approximately orthogonal cross-frequency interactions, reducing overfitting risks.
    
    \item  We validate \oursys's performance by comprehensive experiments, achieving a 7.06\% average improvement in MSE and 3.88\% in MAE across 4 real-world datasets against $10$ well-acknowledged baselines.
    
\end{itemize}

\section{Related Works}
We summarize three key related aspects of existing multivariate time-series research as follows:

\textbf{Time Lag Modeling.} To address temporal misalignment between series, existing methods focus on lag estimation and series alignment. TLGNN\cite{Timelag2} and LagTS \cite{Timelag3} determine optimal lag windows by fitting lag effects through neural networks in the time domain, while LIFT\cite{LIFT} reduces computational complexity of temporal similarity calculation via Pearson coefficient approximation based on Fourier transform. These methods primarily achieve alignment through surface-level similarity matching but fail to capture phase correlations in periodic manufacturing processes. It leads to suboptimal modeling of time-lagged interaction in multi-process manufacturing scenarios.\textbf{(Challenge 1)}.

\textbf{Temporal Pattern Modeling.}
Existing methods employ various model structures to capture temporal patterns in sequences: (1) \textbf{MLP-based methods} \cite{Dlinear, Rlinear} directly map input features to embedding space through linear transformations, showing simple and efficient performance in some scenarios; (2) \textbf{Recurrent models} like LoadDynamics \cite{jayakumar2020self} and TRUST\cite{TKDE24} leverage LSTM structures to capture temporal trends; (3) \textbf{Convolution-based} models such as TimesNet \cite{TimesNet} and MSCNet \cite{TSC2} extract local temporal features via kernel operations; (4) \textbf{Transformers} including KAE-Informer \cite{kaeinformer} and PatchTST \cite{PatchTST} model long-range dependencies through attention mechanisms. Recent approaches \cite{FEDformer,FITS, PDF} decompose series into frequency sub-bands using Fourier or wavelet transforms and model features in the frequency domain. However, these methods either fail to disentangle overlapping periodic components or aggregate heterogeneous frequencies, making them unable to isolate the distinct impacts of multi-scale operations in multi-process manufacturing, thus struggling with compound periodicity \textbf{(Challenge 2)}.

\textbf{Series Dependency Modeling.}
Modeling inter-series dependencies is crucial for multi-process  quality prediction, as it reveals how processes influence each other. Existing methods mainly rely on two frameworks. (1) Graph-based methods: GNNs \cite{MTGNN, STAMP, MAST, MSGNet} construct graphs using statistical correlations or neural networks to model series dependencies; (2) Attention-based methods like \cite{iTransformer, Crossformer, HARMONY} tokenize series and utilize cross-attention to capture global series dependencies. Both approaches fall short in addressing the frequency-dependent nature of manufacturing process relationships, leaving frequency-decoupled dependencies unaddressed, which is critical for accurate quality prediction. \textbf{(Challenge 3)}.

\section{Problem Definition}
We define the core components of manufacturing workflows and then formalize the multi-process quality prediction task.

\begin{myDef}[\textbf{Process}]
A process refers to a continuous manufacturing stage consisting of multiple operations with distinct periodicities. It records quality metrics at each timestamp, forming a time series 
$\boldsymbol{P} \in \mathbb{R}^{N \times T}$ where $K$ is the number of quality metrics, $T$ is the number of timestamps, and 
$\boldsymbol{P}_{n,t}$ denotes the $n$-th quality metric at timestamp $t$.
\label{Def:Process}
\end{myDef}

In practical manufacturing systems, multiple processes are interconnected to form a complete production chain.

\begin{myDef}[\textbf{Multi-Process}]
A multi-process system comprises $M$ interdependent processes  $\{\boldsymbol{P}_1, \boldsymbol{P}_2, \dots, \boldsymbol{P}_M\}$. Their quality metrics collectively form a multivariate time series  $\boldsymbol{X} \in \mathbb{R}^{ M \times T}$ where $\boldsymbol{X}[m,t]$
 represents the quality metric of the $m$-th process at timestamp $t$.
\label{Def:MultiProcess}
\end{myDef}

The task of predicting quality metrics across interconnected processes is defined as follows:
\begin{myDef}[\textbf{Multi-Process Quality Prediction}]
Given the multi-process quality series $\boldsymbol{X} \in \mathbb{R}^{M \times N \times T}$ , our goal is to predict the quality metrics of each process for the next $H$
 timestamps, denoted as $\hat{\boldsymbol{X}} \in \mathbb{R}^{M  \times H}$. where $\hat{\boldsymbol{X}} $
 approximates the ground-truth metrics $\boldsymbol{Y} \in \mathbb{R}^{M  \times H}$.
\label{Tsk:Prediction}
\end{myDef}

\section{Methodology}
\begin{figure*}[htpb]
\centering
\includegraphics[width=1\textwidth]{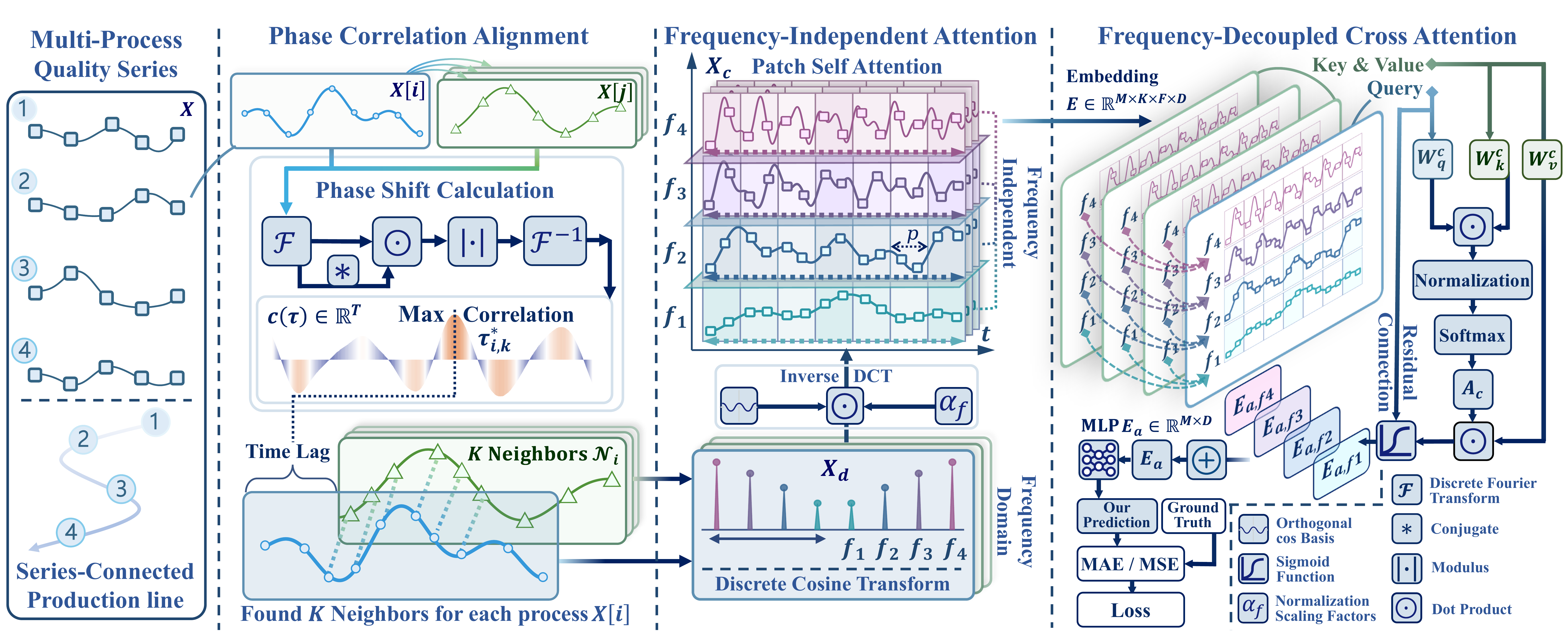}
    \caption{\oursys Framework. For the input Multi-Process Quality Series, we first perform Phase Correlation Alignment to alleviate time lags. Subsequently, patch attention is applied to the independent frequency components of the series. Finally, cross attention is utilized to capture the connections between the same-frequency components of the series for prediction.}
\label{fig:model}
\end{figure*}

 The proposed \oursys is comprised of three main components: (1)  Phase Correlation Alignment for mitigating  time-lags, (2) Frequency-Independent Patch Attention for single-series periodic feature extraction, and (3) Frequency-Decoupled Cross Attention for inter-series dependency modeling.  Fig. ~\ref{fig:model} presents our framework. In the following parts, we introduce each of these three components.

\subsection{Phase Correlation Alignment}
Manufacturing processes with operation-dependent time lags challenges time-domain alignment methods, which prioritize amplitude over phase correlation. We propose a frequency-domain phase approach to address the limitation.

For each target process $i$, we quantify phase coherence with other processes by the phase correlation in Eq.~(\ref{eqn: PC}). 
\begin{equation}\label{eqn: PC}
    \mathcal{R}_{i,j}(\tau) = \mathcal{F}^{-1} \left( \frac{\mathcal{F}(\mathbf{X}[i, :]) \odot \overline{\mathcal{F}(\mathbf{X}[j, :])}}{\left| \mathcal{F}(\mathbf{X}[i, :]) \odot \overline{\mathcal{F}(\mathbf{X}[j, :])} \right|} \right) (\tau).
\end{equation}
$\mathcal{F}$ and $\mathcal{F}^{-1}$ denotes the Discrete Fourier Transform and its inverse.  $\overline{\cdot}$ represents complex conjugation. $\odot$ denotes element-wise multiplication. $\left| \cdot \right|$ is used to normalize out amplitude.
Different from amplitude-based similarity metrics like Pearson Correlation that conflate amplitude variations with temporal alignment, $\mathcal{R}_{i,j}(\tau) \in \mathbb{R}^T$ exclusively measures phase shift between $i$ and $j$ at lag $\tau$, making it robust to amplitude fluctuations from varying production scenarios.

With $\mathcal{R}_{i,j}(\tau)$, we select the top $K$ correlated neighbors for process $i$ as Eq.~(\ref{eqn: FindK}) denotes. $\mathcal{N}_i \in \{1,\dots,M\}^K$ are the indices of the $K$  neighbors for process $i$. 
\begin{equation} \label{eqn: FindK}
     \mathcal{N}_i = \arg\max_{j \in \{1, ..., M\}} \left\{ \max_{\tau} \mathcal{R}_{i,j}(\tau) \right\}_{k=1}^K.
\end{equation} 

We then apply dynamic shifting to variable lags across processes, avoiding the constraints of static lag assumptions.
For each neighbor $k\in \{1,\dots,K\}$, we pick the time lag $\tau_{i,k}^* = \arg\max_{\tau} \mathcal{R}_{i, \mathcal{N}_i[k]}(\tau)$ that maximizes phase alignment between process $i$
 and its $k$-th neighbor. Based on $\tau_{i,k}^*$, we align the $k$-th neighbor to process $i$ as Eq.~(\ref{eqn: align}) describes.
\begin{equation}\label{eqn: align}
    \mathbf{X}_{\text{a}}[i, k, t] = \mathbf{X}[\mathcal{N}_i[k], (t + \tau_{i,k}^*) \mod T].
\end{equation}
The aligned output $\mathbf{X}_{\text{a}} \in \mathbb{R}^{M\times (K+1) \times T}$  resolves time-lagged series interactions by leveraging phase coherence, enabling downstream modules to model sequential interactions. Specifically, $k=0$ denotes the target series itself.

\subsection{Frequency-Independent Patch Attention}
Quality series involve multi-scale periodic operations that generate compound periodic patterns. These patterns are challenging to model when aggregated, as different operations affect quality through distinct periodic mechanisms. We address this by decomposing series into frequency-specific components and modeling each independently.

We first decompose each aligned series into $F$ orthogonal  frequency bands by the Discrete Cosine Transform (DCT)  \cite{DCT} , which reduces artificial high-frequency noise by handling non-periodic boundaries through symmetric extension.  For an aligned series segment $\mathbf{X}_{\text{a}}[i,k,:]$, DCT are performed as Eq.~(\ref{eqn:dct}) denotes. 
\begin{equation}\label{eqn:dct}
\mathbf{X}_{\text{d}}[i, k, f] = \alpha_f \sum_{s=0}^{T-1} \mathbf{X}_{\text{a}}[i, k, s] \cdot \cos\left( \frac{\pi f (2s + 1)}{2T} \right),
\end{equation}
where $\alpha_f= \sqrt{1/T}$ for $f=0$, $\alpha_f= \sqrt{2/T}$  for $f=\{1,\dots,F-1\}$, and  $f\in\{0,\dots,F-1\}$ indexes frequency bands. Each $\mathbf{X}_{\text{d}}[i, k, f]$ captures the $f$-th frequency component, isolating periodic patterns from specific operations.   DCT basis functions \textbf{ensure decomposed frequency components are mutually orthogonal and statistically independent}. This orthogonality is critical for isolating distinct operational periodicities without cross-interference.

To reconstruct the $f$-th frequency-specific temporal pattern, we apply inverse DCT to each frequency in Eq.~(\ref{eqn:idct}).
\begin{equation}\label{eqn:idct}
\mathbf{X}_{\text{c}}[i,k,f,t] = \alpha_f \mathbf{X}_{\text{d}}[i,k,f] \cdot \cos\left( \frac{\pi f (2t + 1)}{2T} \right).
\end{equation}

We then split each frequency component into non-overlapping temporal patches to model local periodic patterns at the granularity of individual operations. For frequency band $f$, $\mathbf{X}_{\text{c}}[i, k, f, :]$ is splited into  $N_p = \lfloor T/P \rfloor + 1$ patches where P denote the patch length as Eq. ~(\ref{eqn:patch}) denotes.
\begin{equation}\label{eqn:patch}
\mathbf{X}_p[i,k,f,b,:] = \mathbf{X}_{\text{c}}[i,k,f, bP: max((b+1)P,t)],
\end{equation}
where $b=1,...,N_p$ indexes patches. To preserve the unique periodicity of each operational scale, we apply attention exclusively within each frequency. For frequency $f$, the attention weights between patches are descirbed in Eq.~(\ref{eqn:attn}).
\begin{equation}\label{eqn:attn}
\mathbf{A}[b_1,b_2] = \delta\left( \frac{(\mathbf{W}_q^f \mathbf{X}_p[b_1,:]) \cdot (\mathbf{W}_k^f \mathbf{X}_p[b_2,:])^T}{\sqrt{D}} \right), 
\end{equation}
 where $\delta$ denotes the softmax function. $\mathbf{W}_q^f, \mathbf{W}_k^f, \mathbf{W}_v^f\in \mathbb{R}^{D \times P}$ are frequency-specific query/key/value projections, and $D$ is the embedding dimension.
The aggregated feature for frequency $f$ is denoted in Eq.~(\ref{eqn:feat}).
\begin{equation}\label{eqn:feat}
\mathbf{E}[i, k, f, :] = \sum_{b=1}^{N_{p}} \mathbf{A}[i, k, f, :, b] \cdot (\mathbf{W}_v^f \mathbf{X}_p[i, k, f, b, :]).\end{equation}
$\mathbf{E}[i, k, f, :]\in \mathbb{R}^{ D}$ encodes the  $f$-th frequency-specific patterns of the $k$-th neighbor of process $i$ with frequency-specific attention ensuring that each operational scale’s impact on quality is modeled independently.

\subsection{Frequency-Decoupled Cross Attention}
\begin{figure}[htbp]
\centering
\includegraphics[width=\linewidth]{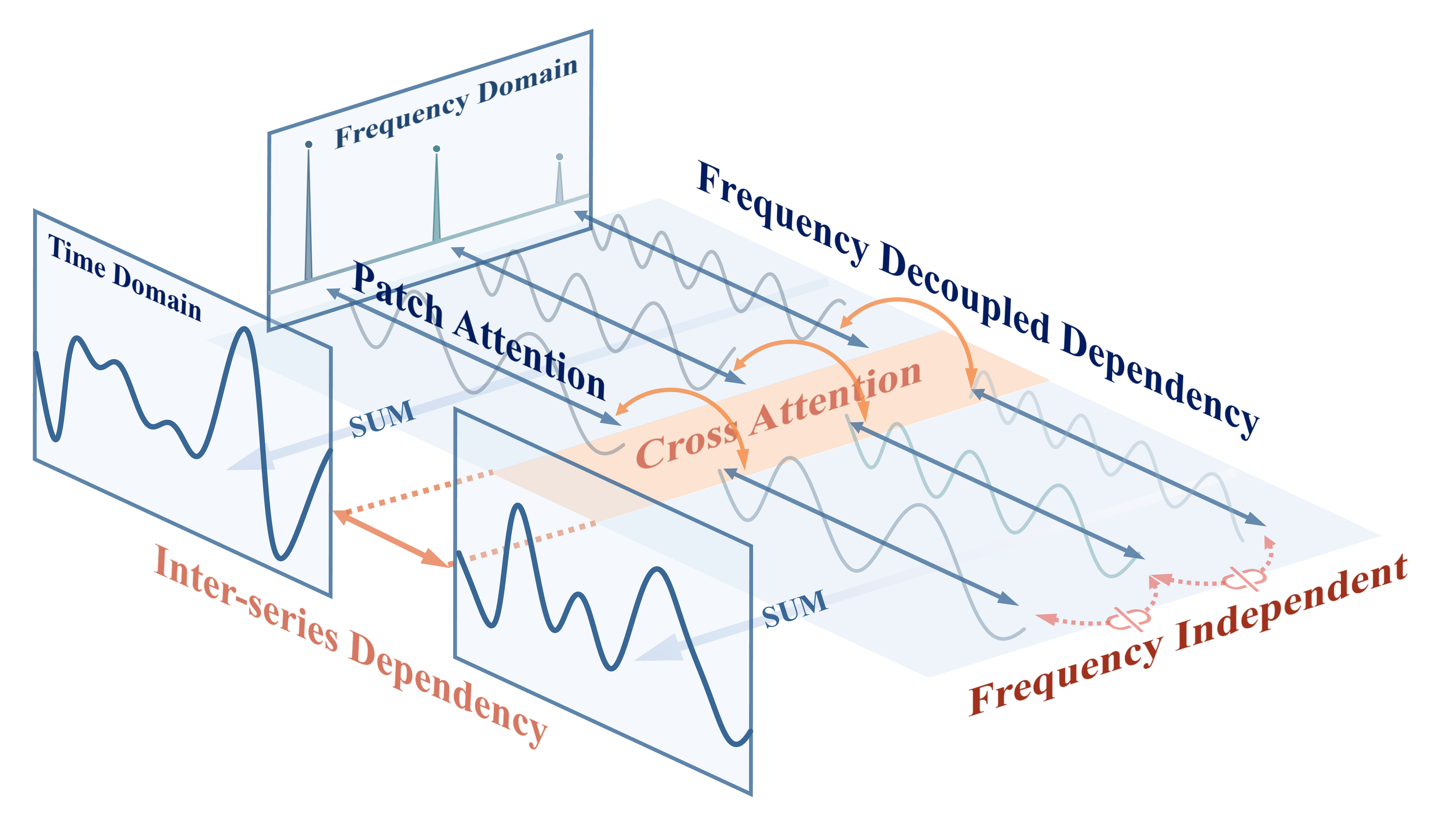}
\caption{Illustration of Frequency-Independent and Decoupled Attention.
Patch attention operates within each independent frequency component of individual series. Cross attention decouples series dependency at the frequency level, modeling relationships between identical frequencies.
}
\label{fig: math}
\end{figure}

Inter-series dependencies in manufacturing are frequency-specific, with meaningful correlations concentrated in shared frequency bands.  We address this by designing a frequency-decoupled cross-attention mechanism that exclusively models dependencies between same-frequency components across different processes. Fig.~\ref{fig: math} illustrates the patch attention and cross attention modules we design.

For a target process $i$, we update its frequency-specific target representation $\mathbf{E}[i, 0, f, :] \in \mathbb{R}^D$ using neighbor features $\mathbf{E}[i, k, f, :] \in \mathbb{R}^D$   from the same frequency band $f$. 

For each frequency $f$,  cross-attention weights between target $i$
 and its $K$ neighbors are computed as Eq.~(\ref{eqn:cross-attn}) denotes.  The query is derived from the target's own frequency-specific feature $\mathbf{E}[i, 0, f, :]$ while keys and values are from its neighbors as Eq.~(\ref{eqn:cross-attn}) denotes.

\begin{equation}\label{eqn:cross-attn}
\mathbf{A}_{\text{c}}[i, f, k] = \delta\left( \frac{(\mathbf{W}_q^c \mathbf{E}[i, 0, f, :]) \cdot (\mathbf{W}_k^c \mathbf{E}[i, k, f, :])^T}{\sqrt{D}} \right),
\end{equation}
where $\mathbf{W}_q^c, \mathbf{W}_k^c \in \mathbb{R}^{D \times D}$ are frequency-specific projection matrices and $k \in \{0, ..., K\}$ indexes processes.  This frequency-decoupled attention ensures only same-frequency components contribute to the target's representation update, avoiding noise from cross-frequency interactions.  The target's representation is then updated by aggregating neighbor information via the computed attention weights as Eq.~(\ref{eqn:cross-update}).
\begin{equation}\label{eqn:cross-update}
\mathbf{E}_a[i, f, :] =  \sum_{k=0}^K \mathbf{A}_{\text{c}}[i, f, k] \cdot (\mathbf{W}_v^c \mathbf{E}[i, k, f, :]) .
\end{equation}

$\mathbf{W}_v^c \in \mathbb{R}^{D \times D}$  is the value projection matrix and $\mathbf{E}_a[i, f, :] \in \mathbb{R}^D$ is the updated representation of the target process $i$ at frequency $f$.  We then aggregate the  all updated frequency-specific representations to form the final target process representation as $\mathbf{E}_{a}[i, :] = \sum_{f=1}^{F} \mathbf{E}[i, f, :] \in \mathbb{R}^{M\times D}$. 

This aggregation preserves multi-scale frequency information while maintaining the integrity of same-frequency dependencies. Finally, we project the aggregated representation to the prediction space by a multi-layer perceptron ($MLP: \mathbb{R}^{M\times D} \rightarrow \mathbb{R}^{M\times H}$) as Eq.~(\ref{eqn: MLP}) denotes.

\begin{equation}\label{eqn: MLP}
\hat{\mathbf{X}} = \text{MLP}(\mathbf{E}_{a}) \in \mathbb{R}^{ H}.
\end{equation}
The model is trained using the mean squared error (MSE) loss between predicted and ground-truth quality metrics as Eq.~(\ref{eqn: loss}) describes.  $\mathbf{Y}[i, h]$ is the ground-truth quality metric. This loss function encourages the model to minimize prediction errors, ensuring accurate quality forecasting.
\begin{equation}\label{eqn: loss}
\mathcal{L} = \frac{1}{M \cdot H} \sum_{i=1}^M \sum_{h=1}^H (\hat{\mathbf{X}}[i, h] - \mathbf{Y}[i, h])^2.
\end{equation}

\section{Experiments}
In this section, we conduct a series of experiments designed to address the following research questions:

\textbf{RQ1:} How does the performance of \oursys compare with SOTA methods in predicting multi-process series?

\textbf{RQ2:} Can the key components within \oursys be identified as significant contributors to its superior performance?

\textbf{RQ3:}  How do different parameter settings impact the predictive performance of \oursys?

\subsection{Experimental Settings}

\begin{table*}[h]
\centering
\caption{Comparative Results for Forecasting Performance. The best results are in boldface and the second best results are underlined. Avg. Ranks refer to the average ranking of the corresponding model across different datasets and prediction horizons. }\label{tab:comp}
\scalebox{0.77}{
\renewcommand{\arraystretch}{1.1} 

\setlength{\tabcolsep}{1.5mm}
\begin{tabular}{cccccccccccccc}
\toprule
Datasets & Metrix & H & Fedformer & Crossformer & PatchTST & iTransformer & SimpleTM & DLinear & NLinear & FITS & SASGNN & LIFT & \oursys \\ \midrule
\multirow{8}{*}{\textbf{MP}} & \multirow{4}{*}{MSE} & 12 & 0.6820 & 0.5210 & 0.4997 & 0.5304 & 0.5384 & 0.8453 & 0.9070 & 0.5375 & \underline{0.4587} & 0.4722 & \textbf{0.4346}\\
 &  & 24 & 0.7974 & 0.6461 & 0.5271 & 0.5555 & 0.5715 & 0.8532 & 0.9855 & 0.5768 & 0.5668 & \underline{0.5138} & \textbf{0.4891}\\
 &  & 36 & 0.8592 & 0.5768 & 0.5414 & 0.5707 & 0.5905 & 0.8459 & 1.0066 & 0.5827 & 0.5563 & \underline{0.5311} & \textbf{0.5074}\\
 &  & 48 & 0.8799 & 0.6115 & 0.5513 & 0.5811 & 0.6055 & 0.8621 & 1.0081 & 0.5807 & 0.6325 & \underline{0.5467} & \textbf{0.5200} \\ \cline{2-14}
 & \multirow{4}{*}{MAE} & 12 & 0.6173 & 0.5356 & 0.5041 & 0.5233 & 0.5287 & 0.4975 & 0.7264 & 0.5387 & 0.4587 & \underline{0.4847} & \textbf{0.4601}\\
 &  & 24 & 0.6701 & 0.6158 & 0.5207 & 0.5374 & 0.5471 & 0.7031 & 0.7560 & 0.5682 & 0.5440 & \underline{0.5090} & \textbf{0.4896}\\
 &  & 36 & 0.6957 & 0.5668 & 0.5281 & 0.5707 & 0.5573 & 0.6975 & 0.7642 & 0.5696 & 0.5454 & \underline{0.5188} & \textbf{0.5010}\\
 &  & 48 & 0.7049 & 0.5869 & 0.5341 & 0.5811 & 0.6055 & 0.7063 & 0.7653 & 0.5807 & 0.6325 & \underline{0.5337} & \textbf{0.5200} \\ \hline
\multirow{8}{*}{\textbf{Production}} & \multirow{4}{*}{MSE} & 12 & 0.4594 & 0.2676 & \underline{0.0283} & 0.0364 & 0.0365 & 0.6253 & 0.0406 & \underline{0.0283} & 0.1056 & 0.0301 & \textbf{0.0202}\\
 &  & 24 & 0.6212 & 0.3400 & \underline{0.0727} & 0.0926 & 0.0904 & 0.6410 & 0.0922 & 0.0797 & 0.2973 & 0.0734 & \textbf{0.0568}\\
 &  & 36 & 0.7300 & 0.3996 & \underline{0.1153} & 0.1715 & 0.1586 & 0.7017 & 0.1649 & 0.1539 & 0.6264 & 0.1342 & \textbf{0.0971}\\
 &  & 48 & 0.8548 & 0.4721 & \underline{0.1794} & 0.3099 & 0.2384 & 0.7786 & 0.2550 & 0.2463 & 0.8584 & 0.2688 & \textbf{0.1741} \\ \cline{2-14}
 & \multirow{4}{*}{MAE} & 12 & 0.3312 & 0.1345 & \underline{0.0663} & 0.0784 & 0.0765 & 0.4147 & 0.0923 & 0.0732 & 0.2044 & 0.0738 & \textbf{0.0594}\\
 &  & 24 & 0.3646 & 0.1867 & \underline{0.1122} & 0.1312 & 0.1226 & 0.4215 & 0.1383 & 0.1306 & 0.3158 & 0.1181 & \textbf{0.0947}\\
 &  & 36 & 0.3969 & 0.2071 & \underline{0.1442} & 0.1788 & 0.1658 & 0.4420 & 0.1835 & 0.1840 & 0.4547 & 0.1571 & \textbf{0.1302}\\
 &  & 48 & 0.4360 & 0.2525 & \underline{0.1832} & 0.2393 & 0.2076 & 0.4714 & 0.2272 & 0.2329 & 0.5030 & 0.2142 & \textbf{0.1817} \\ \hline
\multirow{8}{*}{\textbf{Mining}} & \multirow{4}{*}{MSE} & 12 & 0.4583 & 0.3848 & 0.3651 & 0.3942 & 0.3853 & 0.5677 & 0.4318 & 0.3979 & 0.4915 & \underline{0.3632} & \textbf{0.3617}\\
 &  & 24 & 0.5451 & 0.4664 & \underline{0.4605} & 0.5431 & 0.4698 & 0.5948 & 0.5387 & 0.4809 & 0.6992 & 0.4610 & \textbf{0.4579}\\
 &  & 36 & 0.5445 & 0.4996 & 0.5062 & 0.5494 & 0.5116 & 0.6078 & 0.5809 & 0.5144 & 0.7049 & \underline{0.4983} & \textbf{0.4971}\\
 &  & 48 & 0.5757 & \underline{0.5172} & 0.5252 & 0.5431 & 0.5396 & 0.6181 & 0.6004 & 0.5349 & 0.7529 & 0.5259 & \textbf{0.5116} \\ \cline{2-14}
 & \multirow{4}{*}{MAE} & 12 & 0.3782 & 0.3456 & \underline{0.2951} & 0.3130 & 0.3021 & 0.4455 & 0.3425 & 0.3196 & 0.3862 & 0.3099 & \textbf{0.2904}\\
 &  & 24 & 0.4175 & 0.3908 & 0.3308 & 0.3485 & 0.4674 & 0.5485 & 0.3994 & 0.3919 & 0.3853 & \underline{0.3299} & \textbf{0.3253}\\
 &  & 36 & 0.4193 & 0.4130 & 0.3559 & 0.3807 & 0.3729 & 0.6480 & 0.4233 & 0.3861 & 0.4732 & \underline{0.3536} & \textbf{0.3527}\\
 &  & 48 & 0.4378 & 0.4197 & \underline{0.3794} & 0.3919 & 0.3879 & 0.4774 & 0.4351 & 0.4017 & 0.5030 & 0.3798 & \textbf{0.3759} \\ \hline
\multirow{8}{*}{\textbf{Manufacturing}} & \multirow{4}{*}{MSE} & 12 & 0.9891 & \underline{0.9658} & 1.0306 & 1.0309 & 1.0003 & 1.2283 & 1.3050 & 1.0005 & 1.1577 & 1.0313 & \textbf{0.9649}\\
 &  & 24 & 0.9952 & \underline{0.9695} & 1.0315 & 1.0135 & 0.9884 & 1.2417 & 1.3413 & 1.0008 & 1.3069 & 1.0296 & \textbf{0.9683}\\
 &  & 36 & 1.0517 & \underline{0.9712} & 1.0349 & 0.9897 & 1.0332 & 1.2482 & 1.3308 & 1.0091 & 1.2203 & 1.0345 & \textbf{0.9708}\\
 &  & 48 & 1.0363 & \underline{0.9733} & 1.0383 & 0.9855 & 1.0291 & 1.2505 & 1.3285 & 1.0033 & 1.4767 & 1.0381 & \textbf{0.9720} \\ \cline{2-14}
 & \multirow{4}{*}{MAE} & 12 & \underline{0.7841} & 0.7872 & 0.7984 & 0.8006 & 0.7893 & 0.8777 & 0.8981 & 0.7898 & 0.8521 & 0.7996 & \textbf{0.7779}\\
 &  & 24 & \underline{0.7834} & 0.7888 & 0.7997 & 0.7942 & 0.7846 & 0.8821 & 0.9093 & 0.7884 & 0.8924 & 0.7992 & \textbf{0.7823}\\
 &  & 36 & 0.8083 & 0.7889 & 0.8013 & \underline{0.7854} & 0.8028 & 0.8839 & 0.9070 & 0.7922 & 0.8682 & 0.8010 & \textbf{0.7851}\\
 &  & 48 & 0.8049 & 0.7895 & 0.8026 & \underline{0.7844} & 0.8011 & 0.8850 & 0.9065 & 0.7906 & 0.9577 & 0.8029 & \textbf{0.7826}\\ \hline
\multirow{2}{*}{\textbf{Avg. Rank}} & MSE & \multirow{2}{*}{All} & 8.3125 & 5.0625 & \underline{3.8750} & 5.8750 & 5.3125 & 9.8750 & 9.0000 & 5.0625 & 8.5625 & 4.0000 & \textbf{1.0000}\\
 & MAE &  & 8.1250 & 6.5625 & \underline{3.5625} & 4.9375 & 5.0000 & 9.8125 & 9.0000 & 5.7500 & 8.3125 & 3.9375 & \textbf{1.0000}\\
\bottomrule
\end{tabular}
}

\end{table*}

\subsubsection{Datasets}
 Our study leverages four open-source industrial datasets spanning diverse manufacturing scenarios:
 
\textbf{ (1) MP}\footnote{https://www.kaggle.com/datasets/supergus/multistage-continuousflow-manufacturing-process}: It has 116 processes with 1-second intervals over an 4-hour period, from a continuous flow two-step manufacturing line, where step 1's output is step 2's input. We selected 11 processes, excluding columns with low temporal variability. 
\textbf{ (2)	Production}\footnote{https://www.kaggle.com/datasets/podsyp/production-quality/}: It has 18 processes with 1-hour intervals over a 3-year period, from measurements of temperature sensors from 5 chambers in a roasting machine. We extracted 1-year data for the completeness of production cycle. Each chamber has 3 sensors, representing complex temperature dynamics measured across processes.
\textbf{ (3)	Mining}:\footnote{https://www.kaggle.com/datasets/edumagalhaes/quality-prediction-in-a-mining-process}: It has 24 processes with 20-second intervals over a 6-month period, including sensor data from a mineral flotation process. We extracted a subset  based on the completeness of production cycle and prioritized 8 key processes.
\textbf{ (4)  Manufacturing}\footnote{https://www.kaggle.com/datasets/rukenmissonnier/manufacturing-data-for-polynomial-regression}: It has 6 processes from a simulated production process with engineered parameters, where temperature/pressure interactions drive synthetic quality metrics. 

The datasets are divided into training, validation, and testing sets in a 7:1:2 ratio for robust model evaluation. Table~\ref{Dataset} summarizes the statistics of the four datasets we used.

 \begin{table}[H]
\setlength\tabcolsep{3.2 pt}
\caption{Statistics of the Datasets for Experiments}
 \scalebox{1}{
\begin{tabular}{ccccc}
\toprule Dataset & \makecell[c]{Num of \\ Process }& \makecell[c]{Time \\Stamps } &\makecell[c]{Sampling\\Frequency} & \makecell[c]{ Industrial\\Scenario} \\
\midrule 
MP & 11 & 4377 & 1 Second & Assembly\\
Production & 18 & 10000 & 1 Hour & Roasting\\
Mining  &   8 & 10000 & 20 Second & Flotation \\
Manufacturing & 6 & 3958 & Irregular & Simulation\\
\bottomrule
\end{tabular}
}
\label{Dataset}
\end{table}

\subsubsection{Baselines}

To evaluate the forecasting performance, we compare \oursys with $10$ well acknowledged time series forecasting methods, comprehensively cover the advanced relevant approaches as follows.  (1) Transformer-based methods: \textbf{FEDformer} \cite{FEDformer} integrates Fourier transform for frequency domain analysis. \textbf{Crossformer} \cite{Crossformer} uses cross-scale attention for multi-scale patterns. \textbf{PatchTST} \cite{PatchTST} applies attention on time series patches.
\textbf{iTransformer} \cite{iTransformer} optimizes for multivariate scenarios with improved embedding. \textbf{SimpleTM} \cite{SimpleTM} adopts frequency domain processing in Transformer framework. 
(2) MLP-based methods: \textbf{DLinear/NLinear} \cite{Dlinear} decompose series into trend/residual components. \textbf{FITS} \cite{FITS} combines MLP with frequency domain analysis.
(3) GNN-based method: \textbf{SASGNN} \cite{SASGNN} is a specialized GNN for multi-process manufacturing quality prediction, using graph to capture industrial variable dependencies. 
(4) Time-lag relationship capturing method: \textbf{LIFT} \cite{LIFT} is a plug-and-play approach that uses PatchTST as its backbone for best performance. It is designed to enhance forecasting by capturing locally stationary time-lag relationships between variates.

\subsubsection{Evaluation Metrics}
We evaluate \oursys's point forecasting performance using two standard metrics: Mean Squared Error (MSE) and Mean Absolute Error (MAE).
MSE measures the average squared difference between predictions and ground truth, penalizing larger errors. MAE calculates the average absolute difference, providing outlier-robust performance assessment.
For multivariate outputs, we report averages across all channels, offering an overall assessment of \oursys's performance across all manufacturing processes. Lower values indicate better accuracy.

\subsubsection{Parameter Settings}

\oursys is developed using PyTorch and trained on a single NVIDIA RTX 4090 GPU. We use an Adam \cite{Adam} optimizer with an initial learning rate of 0.001, the batch size is set to $32$. The dimension of the representation $D$ is configured as $64$, the length of each patch is configured as $64$. We conducted a grid search with the number of aligned neighbors \( K \) in \(\{1, 2, 3, 4, 5, 6, 7\}\) and the number of decoupled frequencies \( F \) in \(\{1, 2, 3, 4, 5\}\). All models are trained for $100$ epochs, and the best-performing model on the validation set is utilized for testing.

\subsection{RQ1: Forecasting Performance} \label{sec: comp}

Table \ref{tab:comp} presents the forecasting performance across multiple horizons. Key insights are summarized as follows:

(1) \oursys outperforms baselines consistently across all scenarios, achieving improvements of 7.06\% in MSE and 3.88\% in MAE. Its generalization stems partly from dynamic phase-correlation alignment – a key edge over LIFT. LIFT’s suboptimal results, arising from time-domain correlation computation, validate the effectiveness of this phase-correlation mechanism.  The mechanism resolves time-lagged interactions across scenarios. Paired with frequency-independent, decoupled attention, \oursys effectively tackles core challenges in manufacturing quality prediction.

(2)  Transformer-based methods such as Crossformer and PatchTST consistently outperform MLP-based counterparts like DLinear and NLinear. Notably, structurally simple methods including FITS and SimpleTM gain competitiveness through frequency-domain analysis. This validates the necessity of explicitly modeling periodic components in manufacturing series, aligning with our insight that frequency-specific patterns are critical for quality prediction.

(3) Manufacturing scenario diversity leads to varying model adaptability. FEDformer performs adequately on the Manufacturing dataset but underfits the Production dataset which features complex inter-series dependencies. This underperformance stems from its inability to handle time lags and frequency-specific relationships. Similarly, SASGNN works well on the MP dataset, where process structures are clear, but degrades on Production and Mining. Their graph-based method struggle to model frequency-decoupled dependencies in depth. These inconsistencies highlight that scenario adaptability in manufacturing requires explicit capture of frequency-specific inter-series relationships.

\subsection{RQ2: Ablation Study}
\begin{figure}[htbp]
\centering
\includegraphics[width=\linewidth]{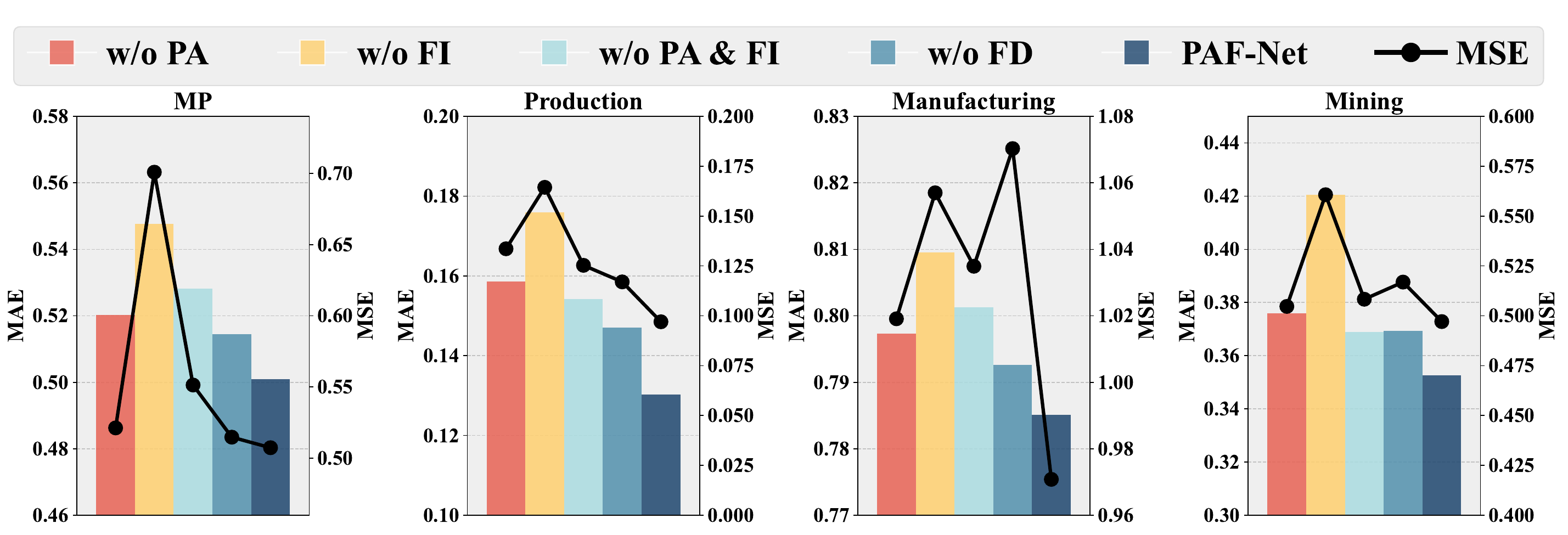}
\caption{Ablation Study. Comparison of forecasting accuracy between \oursys and its variant on MSE and MAE.}
\label{fig:ABStudy}
\end{figure}

We conducted ablation studies to analyze the impact of key components within \oursys, with results shown in Fig. ~\ref{fig:ABStudy}.

(1) Impact of Phase-Correlation Alignment: Removing the phase-correlation alignment module (denoted as ``w/o PA'') leads to consistent performance degradation in both MSE and MAE across all datasets. This deterioration occurs because phase alignment ensures coherent capture of frequency-domain periodic patterns, whose absence disrupts the model’s ability to resolve time-lagged interactions. These results confirm the critical role of frequency-domain alignment in manufacturing scenarios.

(2) Impact of Frequency-Independent Patch Attention: Disabling frequency decomposition and using patch attention directly on raw sequences (``w/o FI'') causes the most significant increases in MSE and MAE. This highlights the importance of decomposing compound periodicities to isolate distinct operational patterns in manufacturing. Notably, further removing phase alignment from this setup (``w/o PA \& FI'') yields performance improvements. This arises because phase alignment, which optimizes frequency-domain coherence, couples inherently with frequency decomposition. Time-domain attention cannot leverage the aligned phase information, making the combined removal less harmful than removing frequency decomposition alone.

(3) Impact of Frequency-Decoupled Cross Attention: Replacing frequency-decoupled cross attention with a mixed-frequency alternative (``w/o FD'') results in consistent MSE and MAE increases. Despite adding model parameters, mixed-frequency modeling ignores the orthogonality of frequency components, introducing spurious relationships during training. This underscores the necessity of explicitly modeling frequency-decoupled inter-series dependencies.

 \subsection{RQ3: Parameter Sensitivity} 
\label{sec: par}

\begin{figure}[htbp]
    \centering  
\includegraphics[width=\linewidth]{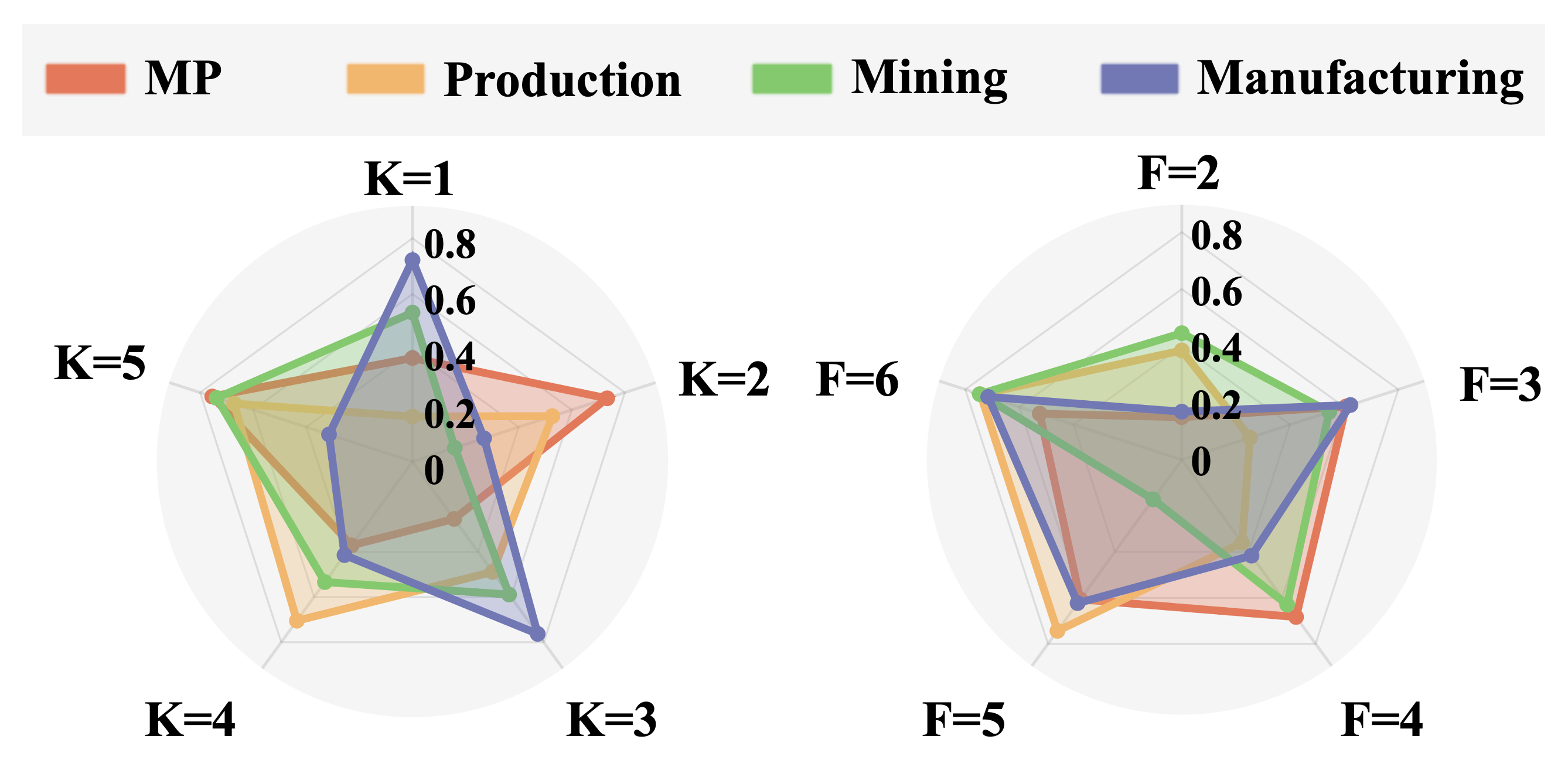}
\caption{Parameter Sensitivity. The two panels respectively show the normalized and mapped MAE results.}
\label{fig:para_dim}
\end{figure}

To explore \oursys’s performance under diverse parameter configurations, we design experiments focusing on two critical  parameters, with results visualized in Fig.~\ref{fig:para_dim}. To account for dataset - scale differences, we apply Z-Score normalization $z=(x-\mu)/\sigma$ to results within the same dataset and then project normalized values to the $(0,1)$ range via the sigmoid function $X=1/(1+e^{-z})$. Therefore, a higher value indicates better performance for a given dataset.


For the number of aligned neighbors $K$, there is no universally optimal value. MP, which features a clear process structure, and Manufacturing, which contains fewer processes, achieve optimal performance when $K=3$ and $2$ respectively. For them, larger $K$ introduce extra noise. In contrast, $K=5$ works better for Production and Mining, which have complex dependencies and more processes, as more neighbors help capture intricate inter-process relationships.

For the number of decoupled orthogonal frequencies $F$, model performance generally improves as more frequencies are decomposed, as exemplified by the Production Dataset. However, this trend is not absolute. MP peaks at $F=4$ and Mining exhibits a significant performance decline when $F=5$. These results indicate $F$ need to be tailored to meet the characteristics of different manufacturing processes.

\section{Conclusion}
We propose \oursys, a specialized time series method designed to tackle time-lagged series interactions, compound periodicities, and frequency-decoupled dependencies in manufacturing processes. Grounded in practical manufacturing scenarios, we conduct an in-depth frequency-domain analysis of in-process periodic operations, underscores the independence of decoupled frequencies, and introduces a novel research perspective. Experimental results on 4 real-world datasets validating \oursys effectiveness.


\nobibliography*

\section{Acknowledgments}
This work was supported by the National Key R\&D Program of China [2024YFF0617700], the National Natural Science Foundation of China [U23A20309, 62172276, 62372296, 62272302], Shanghai Municipal Science and Technology Major Project [2021SHZDZX0102]. 

\bibliography{aaai2026}


\end{document}